\documentclass[11pt]{article}

\usepackage{latexsym}
\usepackage[T1]{fontenc}
\usepackage[latin1]{inputenc}
\usepackage{amsmath,amssymb,amsfonts,nicefrac,microtype}
\usepackage{color,url}
\usepackage{paralist}
\usepackage{dsfont}
\usepackage{graphicx,psfrag}
\usepackage{framed,xcolor,soul}
\usepackage{bbm}
\usepackage{paralist}

\newcommand{\R}{\mathbb{R}}

\newcommand{\Order}{\mathcal{O}}

\newcommand{\one}{\mathbbm{1}}

\begin{document}
\title{Recipe for Fast Large-scale SVM Training:\\Polishing, Parallelism, and more RAM!}

\author{Tobias Glasmachers\\Ruhr-University Bochum, Germany\\\texttt{tobias.glasmachers@ini.rub.de}}

\maketitle

\begin{abstract}
Support vector machines (SVMs) are a standard method in the machine
learning toolbox, in particular for tabular data. Non-linear kernel SVMs
often deliver highly accurate predictors, however, at the cost of long
training times. That problem is aggravated by the exponential growth of
data volumes over time. It was tackled in the past mainly by two types
of techniques: approximate solvers, and parallel GPU implementations. In
this work, we combine both approaches to design an extremely fast dual
SVM solver. We fully exploit the capabilities of modern compute servers:
many-core architectures, multiple high-end GPUs, and large random access
memory. On such a machine, we train a large-margin classifier on the
ImageNet data set in 24 minutes.
%
\end{abstract}

\section{Introduction}

In this paper we pick up a classic learning algorithm, the support
vector machine (SVM). Despite the impressive successes of deep learning
in particular in the areas of image and language processing, there are
still many applications in which the data does not obey a spatial or
temporal structure. Sometimes, data comes as a large table of
unstructured features. This is where classic methods like ensembles and
large margin classifiers shine. Such problems show up regularly in many
application domains like material science, medicine, bioinformatics, and
many more \cite{byvatov2003,lu2013using,ma2014}.

One of the arguably most successful methods for processing tabular data
is the support vector machine (SVM) \cite{cortes1995SVM}. From a modern
perspective, it is limited by one out of two factors. In its linear
form, training is fast, but models are limited to
linear combinations of features, which often precludes accurate
predictors. In its non-linear or kernelized form, high accuracy can be
achieved at the price of long training times, which scale roughly
quadratic with the number $n$ of data points. Since present-day data
sets are orders of magnitude larger than what was common when SVMs were
developed, this is a serious limiting factor for this otherwise highly
valuable method.

The problem of long training times was a very active research topic for
more than a decade, with considerable progress made. The most
influential SVM implementation is for sure the seminal LIBSVM software
\cite{libsvm}. It implements a variant of the sequential minimal
optimization (SMO) method, a dual subspace ascent solver
\cite{osuna1997}. It is complemented by its spin-off LIBLINEAR
\cite{liblinear}, a conceptually similar solver, specialized in training
linear SVMs. These solvers still represent the state-of-the-art for
sequential SVM training of exact solutions on a single CPU core.

Broadly speaking, there are two types of acceleration techniques going
beyond LIBSVM: approximation schemes and parallel algorithms. Most
approximations are based on the insight that restricting the extremely
rich reproducing kernel Hilbert space induced by the kernel to a much
smaller subspace often works well, in the sense that considerable
computational gains can be achieved while sacrificing only very little
predictive performance \cite{tsang2005core,rahimi2008random}. This is
particularly true if the subspace is picked in a data-driven manner
\cite{dekel2007budget,yang2012nystrom}. As an orthogonal development,
parallel algorithms aim to overcome the inherently sequential nature of
subspace ascent, either by employing primal (mini-batch) training
\cite{eigenpro}, or by resorting to heuristics~\cite{thundersvm}.

In this work, we aim to combine both approaches by designing a GPU-ready
dual coordinate ascent algorithm for approximate SVM training. To this
end, we leverage a low-rank approximation technique combined with a dual
linear SVM solver. We add vectorization, multi-core, and GPU processing
to the picture. We achieve considerable speed-ups by paying attention to
the memory organization of our solver. While several SVM solvers
implement a kernel cache, our low-rank technique combined with the large
amount of RAM available in modern server machines allows for a complete
pre-computation of the relevant matrices. We also implement a simplistic
yet robust and effective shrinking method, as well as proper support for
cross-validation and parameter tuning using warm starts. We present an
extremely fast solver suitable for large-scale SVM training on
multi-core systems with and without GPUs.
In a nutshell, our contributions are
\begin{compactitem}
\item[$\bullet$]
	a GPU-ready approximate dual solver for large-scale SVM training in
	minutes (instead of hours or days),
\item[$\bullet$]
	a significantly better compute/memory trade-off through complete
	precomputation of a low-rank matrix factor, and
\item[$\bullet$]
	an robust and effective shrinking technique.
\end{compactitem}

After recalling SVMs in the next section, we summarize existing speed-up
techniques. Then we present our method and its fast implementation. We
demonstrate its power through an empirical comparison with existing
solvers, and finally draw our conclusions.

\section{Support Vector Machines}

{
	\def\OldComma{,}
	\catcode`\,=13
	\def,{%
		\ifmmode%
		\OldComma\discretionary{}{}{}%
		\else%
		\OldComma%
		\fi%
	}%
\paragraph{Primal and Dual Form.~~}
An SVM constructs a decision function of the form%
\footnote{
  The angle brackets denote the inner product in the kernel-induced
  feature space. We drop the bias or offset term \cite{steinwart2011training}.
}
$f(x) \mapsto \langle w, \phi(x) \rangle$. It is directly suitable for
regression tasks, while its sign is considered for binary classification.
Training is based on labeled data
$(x_1, y_1), \dots, (x_n, y_n) \in X \times Y$ and a kernel function
$k : X \times X \to \R$ over the input space $X$. The weight vector
$w^*$ is obtained by solving the (primal) optimization
problem
\begin{align}
	\min_{w \in \mathcal{H}} \quad P(w) = \frac{\lambda}{2} \|w\|^2 + \frac{1}{n} \sum_{i=1}^n \ell\big(y_i, f(x_i)\big), \label{eq:primal}
\end{align}
where $\lambda > 0$ is a regularization parameter, $\ell$ is a loss
function (usually convex in $w$, turning problem~\eqref{eq:primal} into
a convex problem), and $\phi : X \to \mathcal{H}$ is an only implicitly
defined feature map into the reproducing kernel Hilbert space
$\mathcal{H}$, fulfilling $\langle \phi(x), \phi(x') \rangle = k(x, x')$.
The representer theorem allows to restrict the solution to the form
$w = \sum_{i=1}^n \alpha_i y_i \phi(x_i)$ with coefficient vector
$\alpha \in \R^n$, yielding $f(x) = \sum_{i=1}^n \alpha_i y_i k(x, x_i)$.
Training points $x_i$ with non-zero coefficients $\alpha_i \not= 0$ are
called support vectors. For further details we refer the reader to the
excellent review~\cite{bottou2007solvers}.
}

For the simplest case of binary classification (with $Y = \{\pm 1\}$),
the equivalent dual problem \cite{bottou2007solvers} becomes
\begin{align}
	\max_{\alpha \in [0, C]^n} \quad D(\alpha) = \one^T \alpha - \frac{1}{2} \alpha^T Q \alpha. \label{eq:dual}
\end{align}
This is a box-constrained quadratic program, with the notations
$\one = (1, \dots, 1)^T$, $C = \frac{1}{\lambda n}$, and
$Q_{ij} = y_i y_j k(x_i, x_j)$. Corresponding problems for regression
and ranking are of a similar form. Multi-class problems can either be
cast into a larger problem of the same type \cite{mcsvm}, or they are
handled in a one-versus-one manner~\cite{libsvm}.

\paragraph{SVM Training.~~}

Dual decomposition solvers \cite{libsvm,bottou2007solvers} are the
fastest method for training a non-linear SVM to high precision. In each
iteration, they solve a small sub-problem of constant size, which can
amount to coordinate ascent in case of problem~\eqref{eq:dual}.
The sub-problem restricted to a single dual variable $\alpha_i$
(a one-dimensional quadratic program) is solved by the truncated Newton
step
	$\alpha_i \leftarrow \max\big\{0, \min\{C, \alpha_i + \frac{1 - Q_i \alpha}{Q_{ii}} \}\big\},$ 
where $Q_i$ is the $i$-th row of $Q$.
In the simplest case (and in all solvers relevant to
this work), the index $i$ is chosen in a round-robin fashion, possibly
in a randomized order. In elaborate solvers like LIBSVM, considerable
speed-ups can be achieved by a technique called shrinking, which amounts
to temporarily removing variables $\alpha_i$ which remain at the bounds
$0$ or $C$ and hence do not change for a long time. This technique can
reduce the number of active variables to a small subset, in particular
in the late phase of the optimization.

Most alternative solvers operate on the primal
problem~\eqref{eq:primal}. This has advantages and disadvantages. On the
pro side, even simplistic methods like mini-batch stochastic gradient
descent (SGD) \cite{pegasos} can add parallelism in the sense that
multiple data points can enter an update step. This is exploited by the
Eigen-Pro solver \cite{eigenpro}. On the con side, convergence is slow
(although finite-sum acceleration techniques are applicable in some
cases \cite{glasmachers2016finite}), while dual solvers enjoy linear
convergence \cite{lin2001convergence}. Therefore, primal solvers find
rough approximate solutions quickly, while dual methods are the method
of choice when the large margin principle is taken serious, which
requires a rather precise solution.

In any case, the iteration complexity is governed by the computation of
$f(x)$ (or equivalently, by the computation of a partial derivative of
an update of the dual gradient), which is linear in the number of
non-zero coefficients $\alpha_i$. Typically, for each non-zero
coefficient, a kernel computation of cost $\Order(p)$ is required, where
$p$ is the dimension of the input space $X \subset \R^p$, or the average
number of non-zeros for sparse data. The resulting complexity of
$\Order(n p)$ is a limiting factor for large-scale data, since the
number of support vectors grows linearly with
$n$~\cite{steinwart2003sparseness}. In contrast, for linear SVM solvers,
the iteration complexity is simply $\Order(p)$, which is smaller by a
factor of $n$ (often in the order of $10^5$ or more).

\section{Speed-up Techniques}
\label{section:speedup}

\paragraph{Budgeted and Low-Rank Solvers.~~}

The iteration complexity of most primal as well as dual solvers is
tightly coupled to the number of non-zero summands in the weight vector
$w = \sum_{i=1}^n \alpha_i y_i \phi(x_i)$. Most approximate solvers
significantly reduce the number of terms in one way or another.
For a dataset with $n = 10^6$ points, a typical subspace is of dimension
$B=10^3$, where $B$ can be the budget or another parameter controlling
the effective feature space dimension. Such a reduction can be
interpreted as or even constructed through a low-rank approximation of
the kernel matrix~$Q$.

A low-rank method works well if and only if the optimal weight vector
$w^*$ is well preserved by the projection to the low-dimensional
subspace, in the metric induced by $Q$. The fact that low-rank
approximations aim to preserve the eigenspaces corresponding to large
eigenvalues of $Q$ explains why and when such approximations work well,
namely when the spectrum of $Q$ decays sufficiently quickly. For
kernel-induced Gram matrices, this is usually the case, see
\cite{braun2006,eigenpro} and references therein.

Random Fourier Features \cite{rahimi2008random} and related approaches
approximate the most prominent directions of the kernel feature space
$\mathcal{H}$. However, they can be inefficient since the approximation
is performed \textit{a priori}, i.e., without considering the data.
Nyström sampling methods \cite{yang2012nystrom} address this issue by
constructing a data-dependent subspace. In practice, this often amounts
to simply reducing the available basis functions $\phi(x_i)$ to a small
random subset of the training points $\{x_i\}$. Budget methods go even
further by making the subspace adaptive during training
\cite{dekel2007budget,wang2012breaking}. This can be beneficial since in
the end only a single direction needs to be represented in the feature
space, namely the optimal weight vector $w^*$, which is however known
(approximately) only in a late phase of the training process. Budget
methods can be very efficient \cite{qaadan2019}, but they are hard to
parallelize due to their budget maintenance strategy, which usually
amounts to merging support vectors~\cite{wang2012breaking}.

Many different training schemes were designed along these lines. We can
generally differentiate between two-stage methods in which the
construction of the feature space and the training procedure are
separated, and single-stage methods in which the two tasks are performed
simultaneously. This distinction is generally useful, although it
ignores a few variants like lazy learning schemes
\cite{kecman2010locally}. Generally speaking, two-stage methods like
LLSVM \cite{llsvm} put considerable effort into constructing a suitable
subspace and precomputing the projection of the feature vectors into
that subspace. This has the advantage that the second stage reduces to
training a \emph{linear} SVM, which is relatively fast (although with an
iteration complexity of $\Order(B)$ instead of $\Order(p)$, usually with
$B > p$, but still independent of $n$). In contrast, single-stage
approaches save the initial cost and pay the price of non-linear SVM
training, in the hope that the initial saving amortizes. The latter
strategy is generally applied by budget approaches.


\paragraph{GPU-ready Parallel Solvers.~~}

The first GPU-ready SVM solver with a significant impact was ThunderSVM
\cite{thundersvm}. It is a parallel dual subspace ascent method.
Algorithmically, it simply performs the same computations as LIBSVM, but
it executes many subspace ascent steps in parallel. The steps are damped
in order to avoid overshooting, but since there does not seem to be any
rigorous justification (like a convergence proof) for the method, it
should be considered a heuristic. Yet, in practice, it works very
reliably, and it represents considerable progress over LIBSVM's
sequential solver. To the best of our knowledge, it is the only
GPU-ready SVM solver using a dual training scheme.

The EigenPro method \cite{eigenpro} contrasts ThunderSVM im many
aspects. The solver is based on modern deep learning frameworks (there
are Tensorflow and PyTorch versions). Training is based on SGD on the
primal problem. This naturally leverages opportunities for parallelism
by means of mini-batch gradient descent, which comes with convergence
guarantees, although at a slower rate than dual methods. Moreover,
EigenPro is already an approximate solver. It is based on eigen
decomposition of a sub-matrix of $Q$ (based on a random subset of the
data, see Nyström sampling above). On this matrix it performs a
whitening operation. This brings a decisive speedup, since it removes
most of the ill-conditioning from the resulting optimization problem,
which hence becomes much easier to solve with first order methods like
SGD.

\section{A Low-rank Parallel Dual SVM Solver}
\label{section:method}

We aim to combine three types of approaches: (1) we would like to build
on the fast convergence of dual solvers, (2) we would like to achieve
the fast iteration complexity of budget and low-rank methods, and (3) we
aim for an algorithm with GPU-friendly computations. Any combination of
two out of three goals was already demonstrated in the past, see
\cite{qaadan2019,eigenpro,thundersvm}.
In the design of such a method, the following considerations and
trade-offs need to be taken into account:
\begin{itemize}
\item[$\bullet$]
	The matrix $Q$ is of size $n \times n$. For large data sets with
	millions of points, it does not fit into the RAM even of high-end
	server machines. Therefore, existing solvers either work with a
	(row/column-based) kernel cache, or they restrict training to chunks
	of data for a while, before moving on to the next chunk. However
	with a budget $B \ll n$ in place, a factor $G$ of a low-rank
	approximation of $G G^T \approx Q$ is only of size $n \times B$. For
	$B \approx 10^3$ and $n \approx 10^6$, such a matrix fits into the
	available memory of a laptop with 8~GB RAM. For server machines with
	large amounts of RAM, we can even afford two orders of magnitude
	more (e.g., $B \approx 10^4$ and $n \approx 10^7$). At the time of
	writing (March 2022), our largest server machine is equipped with
	512~GB of RAM, while current high-end GPUs come with up to 80~GB of
	RAM. In other words, a complete pre-computation of a low-rank factor
	$G$ becomes feasible even for large%
	\footnote{
	  These are surely not be the largest data sets in existence, but
	  they are definitely large by the standards of the SVM literature.}
	data sets.
\item[$\bullet$]
	As a side effect, whitening the matrix $G G^T$ comes nearly for free
	based on the eigen decomposition of a $B \times B$ sub-matrix of
	$Q$, which is needed anyway for the computation%
	\footnote{
	  A Cholesky decomposition is an attractive alternative at first
	  glance, but since kernel matrices can be ill-conditioned, it
	  regularly runs into numerical problems by requiring \emph{strict}
	  positive definiteness.}
	of the factor $G$. This is akin to the EigenPro method, but less
	relevant in our setting, since the dual coordinate ascent solver is
	not affected by the bad conditioning of the primal problem. On the
	other hand, it is relevant that for some data sets we find a highly
	skewed eigenvalue spectrum.
	As soon as the eigenvalues fall below a threshold close the machine
	precision times the largest eigenvalue, the subspaces are subject to
	strong numerical noise while contributing only minimally to the
	kernel computation. This allows us to drop such components, which
	further reduces the effective dimension (adaptively) and hence
	allows us to process even larger data sets.
\item[$\bullet$]
	Pre-computing the kernel matrix precludes otherwise effective budget
	maintenance techniques like merging of support vectors, since
	merging alters $Q$ (and hence $G$) in a non-linear and
	kernel-dependent fashion. Therefore we settle on a fixed (yet data
	dependent) feature space representation based on a random sample.
	This turns out to be equivalent to the second-most attractive budget
	maintenance strategy: projection onto the remaining support vectors
	\cite{wang2012breaking}. The difference is that all projections are
	pre-computed, hence avoiding an $\Order(B^3)$ operation per SMO
	step.
\item[$\bullet$]
	The precomputation of $G$ effectively turns our approach into a
	two-stage method. Performing a SMO step followed by a
	projection-based budget maintenance operation is exactly equivalent
	to performing a SMO step with an approximate kernel, which is given
	by $G G^T$ instead of $Q$. The second stage indeed reduces to
	solving a linear SVM problem where the original data points are
	replaced with the rows of~$G$.
\item[$\bullet$]
	The excellent study \cite{liblinear} and also our own experience
	clearly indicate that dual methods are generally superior to primal
	SGD-based solvers for obtaining SVM solutions of high quality.
	Therefore we apply a dual coordinate ascent solver, despite the fact
	that it offers fewer opportunities for parallelization. This
	decision is made in the expectation that the faster convergence in
	combination with additional opportunities for parallelization make
	up for it.
\end{itemize}

As indicated above, a large number of approximate SVM training schemes
was already proposed in the literature. Therefore it is not surprising
that we arrived as a solution that is related to existing approaches.
Our solver has conceptual similarity with the low-rank linearization SVM
(LLSVM) \cite{llsvm}, in the sense that it performs the same type of
computations. LLSVM was proposed 10 years ago. Although the two-stage
approach looks quite similar at first glance, there are major and highly
relevant differences:
\begin{compactitem}
\item[$\bullet$]
	LLSVM builds a model based on relatively few but carefully selected
	``landmark'' points, by default 50. However, in our experience, the
	budget size should be in the order of hundreds or better thousands
	in order to achieve a sufficiently good approximation.
\item[$\bullet$]
	LLSVM performs training by iterating over the data set only once,
	where linear SVM training proceeds in chunks of 50,000 points. In
	order to make good use of the precomputed kernel values, 30 epochs
	are performed within each chunk. Hence, each point is used exactly
	30 times, irrespective of the achieved solution accuracy. In
	contrast, in our solver, a standard stopping criterion (similar to
	\cite{liblinear}) is employed to detect convergence.
\item[$\bullet$]
	We remove the concept of chunks altogether and instead demand that
	the full matrix $G$ fits into memory. This leverages the
	capabilities of today's compute servers, simplifies the design, and
	maybe most importantly, it enables fast convergence to the optimal
	solution corresponding to the low-rank kernel represented by~$G$ and
	the application of a corresponding stopping criterion.
\item[$\bullet$]
	We implement all steps of the solver in a multi-core and GPU-ready
	fashion.
\end{compactitem}

\begin{figure}[b]
	\begin{center}
	\psfrag{phase 1}[c][c]{\small phase 1: precomputation of $G$}
	\psfrag{phase 2}[c][c]{\small phase 2: linear training}
	\psfrag{1a}[l][l]{\scalebox{0.7}{\begin{minipage}{52pt}compute the kernel matrix $K$ of size $B \times B$ on a random subset of the data\end{minipage}}}
	\psfrag{1b}[l][l]{\scalebox{0.7}{\begin{minipage}{52pt}compute the eigen decomposition $K = U D U^T$\end{minipage}}}
	\psfrag{1c}[l][l]{\scalebox{0.7}{\begin{minipage}{52pt}compute the factor $L = U \sqrt{D^{-1}}$ of the inverse kernel matrix\end{minipage}}}
	\psfrag{1d}[l][l]{\scalebox{0.7}{\begin{minipage}{52pt}compute the kernel matrix $Z$ between random subset and all points; set $G = Z L$\end{minipage}}}
	\psfrag{2a}[l][l]{\scalebox{0.7}{\begin{minipage}{52pt}compute $f(x_i) = G_i w$\end{minipage}}}
	\psfrag{2b}[l][l]{\scalebox{0.7}{\begin{minipage}{52pt}update $\alpha_i$ (truncated Newton step)\end{minipage}}}
	\includegraphics[width=\textwidth]{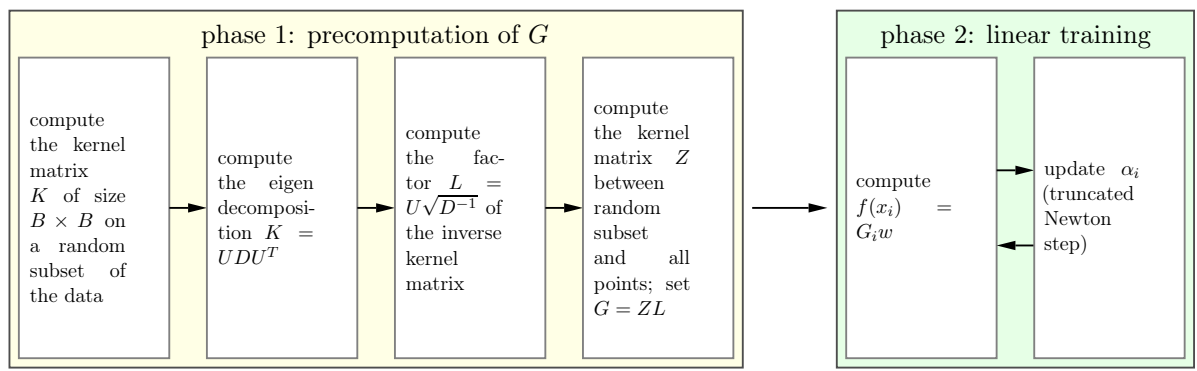}
	\end{center}
	\caption{
		\label{figure:algo}
		Algorithm design as a two-stage method.
	}
\end{figure}

\paragraph{Algorithm Overview.~~}
The resulting algorithmic steps with their two-stage organization are
depicted in figure~\ref{figure:algo}. We call the resulting algorithm
\emph{low-rank parallel dual (LPD)} SVM. For a GPU implementation, it is
important to design all processing steps in such a way that they work in
a streaming fashion, at least for cases in which $G$ fits into CPU
memory but not into GPU memory (this is similar to but not as
restrictive as the chunking approach of LLSVM). The ability to split all
computations into smaller chunks is also important when leveraging
multi-GPU systems.

The multi-stage approach allows for considerable freedom. To this end,
both stages (computation of $G$ and linear SVM training) can be
performed by CPU and GPU, at the discretion of the user. This also holds
for the prediction step (not shown), which is absent in training, but
active in the prediction, test, and cross-validation modes of our
software.

\paragraph{Shinking.~~}
LIBSVM pays considerable attention to ``details'' like efficient caching
and shrinking. Due to complete precomputation of $G$, we do not have a
need for a kernel cache. However, shrinking plays an important role. The
heuristic implemented in LIBSVM is known to be somewhat brittle, but it
can be very effective when working well. We opt for a simplistic
strategy: if a variable was not changed for $k$ steps in a row (we use
$k=5$) then we remove the variable from the problem, and we dedicate a
fixed fraction $\eta$ of the total computation time (say, $\eta = 5\%$)
to checking whether removed variables should be reconsidered. This
heuristic turns out to be far more robust than LIBSVM's strategy which
lacks a systematic way of re-activating variables. We will see in the
next section that shrinking considerably accelerates the linear SVM
training phase.

\paragraph{Cross Validation, Parameter Tuning, and Multi-Class Training.~~}
In reality, we rarely train a single SVM. Kernel SVMs have parameters
like the kernel bandwidth and the regularization parameter $C$
(or $\lambda$) which absolutely need tuning in order to deliver top
performance. While powerful parameter tuning procedures based on
(Bayesian) optimization are widely available, for low-dimensional
problems like this, a simple grid search does the job. It brings the
additional benefit that multiple SVMs are trained with the same kernel,
which allows to reuse the matrix $G$ and hence the complete first stage
of the solver. The same applies to cross-validation: we simply fix the
feature space representation once for the whole data set, pre-compute
$G$, and only then sub-divide the data into folds.%
\footnote{
  While this proceeding may yield a slightly optimistic bias (since some
  basis vectors may stem from the validation set), it is perfectly
  suitable for parameter tuning (since all parameter settings profit in
  the same way), and offers a considerable computational advantage.}
Also, when searching a grid of growing values of $C$, we warm-start the
solver from the optimal solution of the nearest value of $C$ already
completed. None of these techniques is novel, but they are rarely
implemented, although they yield considerable speed-ups. In our solver
their role goes well beyond pure convenience, since many linear training
runs can share the first stage of the computation.

A further important point is the handling of classification problem with
more than two classes. In this regard we follow the design of LIBSVM
\cite{libsvm}, which implements a one-versus-one approach. We do this
for the following reason: one-versus one means that independent SVMs are
trained to separate each pair of classes. What sounds like an immense
burden at first, since the number of pairs grows quadratically with the
number of classes, is more than alleviated by the fact that the
sub-problems are relatively small. Moreover, creating independent
sub-problems is a welcome opportunity for parallelization. For an
in-depth discussion of the pros and cons of this approach we refer the
interested reader to \cite{dogan2016unified}.
In our experiments we demonstrate high efficiency of this scheme on
problems with up to 1000 classes, corresponding to roughly half a
million pairs of classes.

\paragraph{Multi-core and GPU Implementation.~~}
Our implementation uses parallelism at three levels: vectorization
within each single CPU core (implemented by the highly optimized Eigen
C++ library\footnote{\url{https://eigen.tuxfamily.org/}}), multi-core
parallelism (using OpenMP\footnote{\url{https://www.openmp.org/}}),
and GPU parallelism (using CUDA\footnote{\url{https://developer.nvidia.com/}}).
For the latter technology, we distinguish between a single streaming
multiprocessor (SM) and a full GPU. We even implement multi-GPU support
for problems where it makes sense, i.e., if there is sufficient
opportunity for parallelism.

In the first stage of computing $G$, the computationally heavy steps are
batch kernel computation, eigen decomposition, and matrix-matrix
products. All of these are extremely efficient on the GPU, using our own
CUDA kernels, the {cuSOLVER} library, and the {cuBLAS} library,
respectively. On the CPU, we rely on Eigen, and we distribute chunks to
multiple CPU cores using OpenMP. For all of these steps, the GPU turns
out to be far superior to the CPU.

Sadly, ThunderSVM and EigenPro seem to lack proper support for sparse
data, which is commonly encountered in the SVM context. For EigenPro
(relying on deep learning software backends) this is understandable,
since the focus is on images and similar data, which is usually dense.
However, also ThunderSVM converts data to a dense format for kernel
computations. The simple reason is that for most general-purpose kernels
in common use (polynomial, Gaussian and hyperbolic tangent), batch
kernel operations require a matrix-matrix multiplication at their core.
It seems that only specific cases of such products are implemented in
existing GPU libraries, including {cuSPARSE}. In our solver, we
implemented all kernel operations based on efficient sparse matrix
products. For CPU-based processing we rely on the Eigen C++ library,
while for (far more efficient) GPU-based processing, we implemented
sparse matrix products as custom CUDA kernels.

In the second stage of linear SVM training, the trade-offs turn out to
be quite different. In contrast to ThunderSVM, all kernels are
precomputed in the first stage. This is a logical and efficient step,
but it removes most opportunities for parallelism. While ThunderSVM
needs to process $n \cdot p$ floating point values per data points
(where $p$ is the dimension of the input space $X \subset \R^p$), we
only need to handle $B$ values. Although removing opportunities for
parallelism, this is still a great acceleration, since the old wisdom
that the fastest computation is the one that does not run at all still
holds.

As a result, a single SMO (coordinate ascent) step runs extremely fast
on a single CPU in a fully vectorized fashion. The corresponding GPU
implementation is somewhat slowed down by the need for a parallel
reduction operation. Also, the relatively low amount of parallelism
precludes the use of more than one SM on the GPU, since a single SM can
maintain the current weight vector in its fast scratchpad memory, while
the communication cost of multiple SMs would slow down the extremely
fast SMO loop. For current CUDA architectures this means that a total of
``only'' 1024 threads cooperate by computing $B$ summands per data
point or dual variable. To make the orders of magnitude clear: in our
solver, for a realistic value like $B=10^3$, each CPU core performs
several \emph{million} coordinate ascent steps per second. At that pace,
multi-core communication would incur an unacceptable overhead, and the
same holds when communicating between multiple SMs within a GPU.

This leaves us with the (luxury) problem of fully leveraging server
GPUs, which come with more than 100 SMs each. With a single training
run of a binary classification problem this is simply not possible, and
CPU-based training should be preferred. However, when performing even
only a tiny $5 \times 5$ grid search using $10$-fold cross validation on
a 10-class dataset like MNIST, we need to train a total of $5 \times 5
\times 10 \times \binom{10}{2} = 11,250$ binary SVMs: one for each
parameter vector, one for each hold-out set, and one for each pair of
classes. This is far more parallelism than we need to fully exploit even
multiple GPUs at the same time, while the first stage needs to run only
five times (once for each kernel parameter).

Making predictions is relatively fast compared with training. One
decisive difference is that predictions can be computed in parallel, so
that for this task, the GPU is vastly superior to the CPU. Our
implementation therefore defaults to the following behavior: computation
of $G$ and prediction are performed by the GPU and SMO training runs on
the CPU. This is confirmed in the next section as a solid default.

\paragraph{Recipe.~~}

In summary, our approach has conceptually close predecessors in the
existing literature, which we polish up far beyond aesthetics by adding
shrinking, a meaningful stopping criterion, warm starts, and proper
support for cross-validation and parameter tuning. We put considerable
effort into exploiting parallelism in all stages of the solver on CPU
and GPU. Finally, we arrive at quite different trade-offs than older
solvers through a combination of large available memory and memory
saving techniques, in particular the low-rank approach, and by
discarding small eigen values. This is how we arrive at the recipe
\emph{``polishing, parallelism, and more RAM''}
for fast large-scale SVM training.

\section{Experimental Evaluation}

The main evidence for the value of our methodology is of empirical
nature. We present experimental results to answer the following research
questions:
\begin{compactitem}
\item[$\bullet$]
	How fast is our approach compared with existing solvers?
\item[$\bullet$]
	How scalable is our approach, in particular for multi-class problems?
\item[$\bullet$]
	How do the various computational components perform on multi-core
	CPUs and on high-end GPUs?
\item[$\bullet$]
	How does shrinking impact performance?
\end{compactitem}

\begin{table}
	\begin{center}
	{\renewcommand{\arraystretch}{1.2}
	\begin{tabular}{|@{~}l@{~}|@{~}r@{~}|@{~}r@{~}|@{~}r@{~}|@{~}r@{~}|@{~}c@{~}|@{~}l@{~}|}
		\hline
		\textbf{data set} & \textbf{file size} & \# \textbf{classes} & \textbf{size} $n$ & \textbf{budget} $B$ & \textbf{regular.} $C$ & \textbf{kernel} $\gamma$ \\
		\hline
		Adult (a9a) & 2.3~MB &    2 &    32,561 &  1,000 & $2^5$ & $2^{-7}$ \\
		Epsilon     & 12~GB  &    2 &   400,000 & 10,000 & $2^5$ & $2^{-4}$ \\
		SUSY        & 2.4~GB &    2 & 5,000,000 &  1,000 & $2^5$ & $2^{-7}$ \\
		MNIST-8M    & 12~GB  &   10 & 8,100,000 & 10,000 & $2^5$ & $2^{-22}$ \\
		ImageNet    & 59~GB  & 1000 & 1,281,167 &  1,000 & $2^4$ & $2^{-24}$\\
		\hline
	\end{tabular}
	}
	\end{center}
	\caption{
		\label{table:datasets}
		Data sets used in this study, including tuned hyperparameters.
	}
\end{table}

\paragraph{Experimental Setup.~~}
We trained SVMs on one mid-sized and four large data sets, see
table~\ref{table:datasets}. The first four data sets are available from
the LIBSVM website.%
\footnote{\url{https://www.csie.ntu.edu.tw/~cjlin/libsvmtools/datasets/}}
The ImageNet data set \cite{imagenet} can be obtained for free for
non-commercial research.%
\footnote{\url{https://www.image-net.org/}}
We turned it into an SVM training problem by propagating the images
through a pre-trained VGG-16 network (shipped with keras) and extracting
the activations of the last convolution layer, which is of dimension
25,088. The resulting feature vectors are sparse due to the ReLU
activation function.

We used the Gaussian kernel $k(x, x') = \exp(-\gamma \|x-x'\|^2)$ in all
experiments. The hyperparameters $C$ and $\gamma$ were tuned with
grid-search and cross-validation.

All timings were measured on a compute server with 512~GB of RAM, two
Intel Xeon 4216 CPUs (up to 64 concurrent threads), and four NVIDIA A100
GPUs (40~GB RAM each).

In the spirit of open and reproducible research, our software is
available under a permissive open-source BSD-3-clause license.%
\footnote{\url{https://github.com/TGlas/LPD-SVM}}

\begin{table}
	\begin{center}
	{\renewcommand{\arraystretch}{1.2}
	\begin{tabular}{|@{~}l@{~}|@{~}l@{~}|@{~}r@{~}|@{~}r@{~}|@{~}r@{~}|@{~}r@{~}|@{~}r@{~}|}
		\hline
		\textbf{solver} & \textbf{indicator} & \textbf{Adult} & \textbf{Epsilon} & \textbf{SUSY} & \textbf{MNIST-8M} & \textbf{ImageNet} \\
		\hline
		LLSVM      & training   & $1.51$  & $48.38$  & $71.93$  & ---     & --- \\
		           & prediction & $0.25$  & $23.84$  & $29.98$  & ---     & --- \\
		           & error      & $27.3$  & $50.0$   & $27.52$  & ---     & --- \\
		\hline
		ThunderSVM & training   & $2.25$  & $5,315$  & $14,604$ & $7,517$ & $>42$ hours \\
		           & prediction & $1.42$  & $470.51$ & $5,128$  & $11.07$ & --- \\
		           & error      & $14.92$ & $8.70$   & $19.99$  & $0.95$  & --- \\
		\hline
		LPD-SVM    & training   & $2.11$  & $89.86$  & $197.64$ & $868$   & $1,402.86$ \\
		           & prediction & $1.62$  & $12.94$  & $1.22$   & $2.08$  & $36.22$ \\
		           & error      & $14.77$ & $9.85$   & $20.08$  & $1.20$  & $37.52$ \\
		\hline
	\end{tabular}
	}
	\end{center}
	\caption{
		\label{table:performance}
		Performance results of the different solvers: training and prediction time (seconds),
		and classification error (in percent). LLSVM is not applicable to data sets with more
		than two classes. ThunderSVM training on ImageNet stopped after 42 hours with an
		out-of-memory error. At that point, training was about 83\% complete. For LLSVM, we
		observed huge random deviations of predictive performance (about $\pm 13.6\%$ for
		Adult and about $4.6\%$ for SUSY). In the table, we report mean values. Also the best
		values are not competitive compared with the other solvers.
	}
\end{table}

\paragraph{Comparison with Existing Solvers.~~}

\begin{figure}
	\begin{center}
	\includegraphics[width=0.8\textwidth]{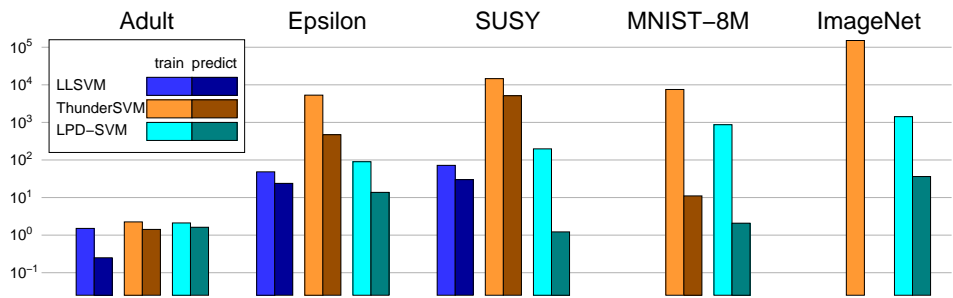}
	\end{center}
	\caption{
		\label{figure:performance}
		Plot of the timing data from table~\ref{table:performance} on a logarithmic scale.
	}
\end{figure}

We aimed to compare with the following solvers:
ThunderSVM,\footnote{\url{https://github.com/Xtra-Computing/thundersvm}}
EigenPro,\footnote{\url{https://github.com/EigenPro}}
and LLSVM.\footnote{\url{https://github.com/djurikom/BudgetedSVM}}
However, we failed to make (the PyTorch version of) EigenPro deliver
meaningful results beyond the built-in MNIST example. On the chosen
data sets, it seems to be quite sensitive data scaling, resulting in
numerical instability. Despite our disappointment, we had to exclude
EigenPro from the comparison. We therefore present performance data for
ThunderSVM, LLSVM, and our own method LPD-SVM in
table~\ref{table:performance}. The same data is displayed graphically
in figure~\ref{figure:performance}.

First of all, while being quite fast (provided that only a single CPU
core is used), LLSVM fails to deliver sufficiently accurate solutions.
On the Epsilon problem, it consistently yields guessing accuracy (50\%
error). This is for its non-adaptive stopping rule, which does not check
for convergence. It is of course easy to be fast if the job is not
complete. This seems to be the case for LLSVM.

In terms of accuracy, LPD-SVM comes quite close to the (nearly exact)
solutions obtained by ThunderSVM. On the Epsilon problem, we pay for the
budget approximation with an increase of the error of a bit more than
one percent, while in the other cases, error rates are quite close. This
slight loss of predictive performance is expected and in line with the
general experience with low-rank approximations in the SVM literature.

At the same time, LPD-SVM systematically outperforms ThunderSVM in terms
of speed, both at training and at prediction time. This is particularly
true for the large data sets where the low-rank approach makes a
decisive difference by lowering the iteration complexity considerably.
We see experimentally that the net difference is roughly one to two
orders of magnitude. For the simple scenario of training a single SVM
classifier, the GPU cannot be fully utilized during training. Hence, the
speed-up over ThunderSVM is fully attributed to the low-rank approach.

\paragraph{Multi-Class SVM Training.~~}

We did not achieve a competitive error rate on the ImageNet problem (the
VGG-16 network performs better). Here, we are mostly interested in
investigating the scaling of our solver to problems with many classes.
Indeed, it takes LPD-SVM only about 24 minutes to train nearly half a
million large-margin classifiers. This corresponds to less than $3$
milliseconds per binary problem.
It seems safe to conclude that---at least computationally---the
one-versus-one approach is very well suited for training large-scale
SVMs.

\paragraph{CPU vs.\ GPU Performance.~~}

When looking at the specifications of modern server CPUs and GPUs then
one would expect the GPU to clearly dominate. However, despite our
effort of implementing an efficient SMO solver natively on the GPU with
custom CUDA kernels, CPUs are still a better match for the inherently
sequential dual algorithm than GPUs. In our empirical data, the CPU wins
the race on both data sets where $G$ fits into GPU memory, and in the
other three cases GPU-based training is not an option at all. The GPU
cannot play its strengths for the simple reason that the SMO loop is
memory-bound, not compute-bound (it is dominated by computing inner
products of vectors of dimension $B$, which is a far too low dimension
for the GPU). While the GPU has a larger net memory throughput, the CPU
has a faster clock speed, more efficient caches, and a very effective
pre-fetching mechanism.

\begin{figure}
	\begin{center}
	\includegraphics[width=0.8\textwidth]{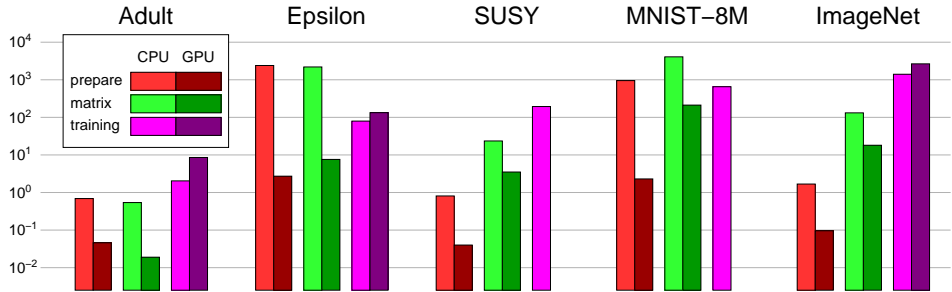}
	\end{center}
	\caption{
		\label{figure:breakdown}
		Timing breakdown into the three stages ``preparation'' (red;
		first three steps in figure~\ref{figure:algo}), ``computation of
		the matrix $G$'' (green; step four in figure~\ref{figure:algo}),
		and ``linear SVM training'' (magenta), on CPU (bright) and GPU
		(dark). All durations are in seconds, depicted on a logarithmic
		scale. For two data sets, training times on the GPU are
		missing since the matrix $G$ does not fit into GPU memory.
	}
\end{figure}

That being said, the overall role of the GPU must not be
under-estimated. It vastly reduces the cost for preparing the matrix
$G$, which can otherwise be the dominant cost on the CPU, and it also
speeds up predictions. This way, the two types of processors (CPU and
GPU) contribute jointly to fast training. If any of the two is dropped
then the overall training times increase significantly.

Figure~\ref{figure:breakdown} gives a more detailed picture by breaking
down the timings of the different stages for GPU and CPU. It becomes
clear that the GPU is far more suitable for the preparation of $G$ and
for making predictions, while the CPU outperforms the GPU for SMO
training. This holds true even for the Imagenet problem where all
$4 \times 108 = 432$ GPU processors are running SMO loops in parallel,
as compared with $64$ CPU cores.

It can also be observed that the trade-offs presented by the two-stage
approach differ quite significantly between data sets, and depending on
whether computations are carried out on CPU or GPU. When using the GPU
then the first stage is generally faster than the second stage, which
means that the investment made in the first stage amortizes. However,
when solving the Epsilon problem completely on the CPU, this is not so
clear, since the computation of $G$ takes the lion's share of the time.

\paragraph{Shrinking.~~}

We evaluate our shrinking algorithm by simply turning it on or off and
measuring the optimization time. In order to achieve clean results, we
restrict time measurements to the second phase (SMO training). Based on
prior experience, we expect shrinking to yield a speed-up in general,
but on the other hand, wrong shrinking decision can cost performance. It
turns out that shrinking is a complete game-changer: without shrinking,
the training time for the Adult data increases by factor 220, while for
Epsilon it increases by factor of 350. Due to excessive training times,
we did not perform the test for all data sets. This impressive speed-up
is in part due to the fact that after removing many variable for fine
tuning in the late phase of the optimization, at memory demand for the
relevant sub-matrix of $G$ reduces and the processor cache becomes for
effective.

\paragraph{Parameter Tuning and Cross-Validation.~~}

We tested the ability of our solver to support hyperparameter grid
search and cross-validation as follows. For $\log_2(C)$, we defined the
grid $\{0, 1, \dots, 9\}$, and we varied $\log_2(\gamma)$ in the range
$\{g^*-2, g^*-1, g^*, g^*+1, g^*+2\}$, where $g^* = \log_2(\gamma^*)$ is
the optimal value, see the rightmost column of table~\ref{table:datasets}.
For each hyperparameter setting, we performed 5-fold cross validation.
Hence, we trained a total of $N = 250 \cdot \binom{c}{2}$ binary SVMs,
where $c$ is the number of classes of the problem. The results are found
in table~\ref{table:multiple}. It lists the total time, the time per
binary problem (the first value divided by $N$), and the speed-up, which
is estimated from the training time from table~\ref{table:performance}
divided by the time per binary problem.

\begin{table}
	\begin{center}
	{\renewcommand{\arraystretch}{1.2}
	\begin{tabular}{|@{~}l@{~}|@{~}r@{~}|@{~}r@{~}|@{~}r@{~}|@{~}r@{~}|}
		\hline
		\textbf{} & \textbf{Adult} & \textbf{Epsilon} & \textbf{SUSY} & \textbf{MNIST-8M} \\
		\hline
		total time              & $247.43$     & $2,837$      & $28,163$      & $84,600$     \\
		time per binary problem & $0.99$       & $11.34$      & $112.65$      & $7.52$       \\
		speed-up                & $\times 2.1$ & $\times 7.3$ & $\times 1.75$ & $\times 2.6$ \\
		\hline
	\end{tabular}
	}
	\end{center}
	\caption{
		\label{table:multiple}
		Timings of the hyperparameter search and cross-validation
		experiments. All times are in seconds.
	}
\end{table}

The speed-up is around a factor of two in most cases, and more than
seven for the Epsilon data. This is in part because the first stage
needs to run only five times (once for each value of $\gamma$), while
previous computations can be reused for the remaining $N-5$ training
runs. This reuse of the precomputation of $G$ is only responsible for
one part of the speed-up, while another part comes from warm-starts
with solutions corresponding to smaller values of~$C$, and from a
better utilization of the compute resources (cores).

\paragraph{Discussion.~~}

Our experimental findings answer our research questions as follows. The
proposed LPD-SVM is extremely fast, with a speed-up of roughly one to
two orders of magnitude over ThunderSVM on large problems. This is
despite the fact that the SMO loop runs on the CPU. LPD-SVM suffers only
a minimal increase of error rates due to the budget approach. It scales
very well to multi-class problems with many classes. We see performance
gains from properly implemented cross-validation and grid search, and
our shrinking algorithm turns out to be extremely effective. We believe
that these results are a valuable addition to the state of the art, and
that our solver provides an interesting alternative existing SVM
software.

\section{Conclusion}

We have presented a GPU-ready SVM solver optimized for compute servers,
called \emph{Low-rank Parallel Dual} (LPD) SVM. It fits into the common
framework of low-rank or budgeted schemes for approximate SVM training.
It takes the typical capabilities of modern server machines into
account: large random access memory, many-core CPUs, and high-end server
GPUs. We demonstrate its potential on a number of large-scale data sets,
achieving state-of-the-art results.

\subsubsection{Acknowledgements}
This work was supported by the Deutsche Forschungsgemeinschaft under
grant number GL 839/7-1.


\begin{thebibliography}{10}
\providecommand{\url}[1]{\texttt{#1}}
\providecommand{\urlprefix}{URL }
\providecommand{\doi}[1]{https://doi.org/#1}

\bibitem{bottou2007solvers}
Bottou, L., Lin, C.J.: Support vector machine solvers (2006)

\bibitem{braun2006}
Braun, M.L.: Accurate error bounds for the eigenvalues of the kernel matrix.
  The Journal of Machine Learning Research  \textbf{7},  2303--2328 (2006)

\bibitem{byvatov2003}
Byvatov, E., Schneider, G.: Support vector machine applications in
  bioinformatics. Applied bioinformatics  \textbf{2}(2),  67--77 (2003)

\bibitem{libsvm}
Chang, C.C., Lin, C.J.: {LIBSVM}: A library for support vector machines. ACM
  Trans. Intell. Syst. Technol.  \textbf{2}(3) (2011)

\bibitem{cortes1995SVM}
Cortes, C., Vapnik, V.: Support-vector networks. Machine learning
  \textbf{20}(3),  273--297 (1995)

\bibitem{dekel2007budget}
Dekel, O., Singer, Y.: Support vector machines on a budget. MIT Press (2007)

\bibitem{imagenet}
Deng, J., Dong, W., Socher, R., Li, L.J., Li, K., Fei-Fei, L.: {ImageNet}: A
  large-scale hierarchical image database. In: IEEE conference on computer
  vision and pattern recognition (CVPR). pp. 248--255 (2009)

\bibitem{dogan2016unified}
Do{\u{g}}an, {\"Ur\"un}., Glasmachers, T., Igel, C.: A unified view on
  multi-class support vector classification. Journal of Machine Learning
  Research  \textbf{17}(45),  1--32 (2016)

\bibitem{mcsvm}
Do\u{g}an, {\"{U}}., Glasmachers, T., Igel, C.: A unified view on multi-class
  support vector classification. Journal of Machine Learning Research (JMLR)
  \textbf{17}(45),  1--32 (2016)

\bibitem{liblinear}
Fan, R.E., Chang, K.W., Hsieh, C.J., Wang, X.R., Lin, C.J.: {LIBLINEAR}: A
  library for large linear classification. J. Mach. Learn. Res. pp. 1871--1874
  (2008)

\bibitem{glasmachers2016finite}
Glasmachers, T.: Finite sum acceleration vs. adaptive learning rates for the
  training of kernel machines on a budget. In: NIPS workshop on Optimization
  for Machine Learning (2016)

\bibitem{kecman2010locally}
Kecman, V., Brooks, J.P.: Locally linear support vector machines and other
  local models. In: The 2010 International Joint Conference on Neural Networks
  (IJCNN). pp.~1--6. IEEE (2010)

\bibitem{lin2001convergence}
Lin, C.J.: On the convergence of the decomposition method for support vector
  machines. IEEE Transactions on Neural Networks  \textbf{12}(6),  1288--1298
  (2001)

\bibitem{lu2013using}
Lu, W.C., Ji, X.B., Li, M.J., Liu, L., Yue, B.H., Zhang, L.M.: Using support
  vector machine for materials design. Advances in Manufacturing
  \textbf{1}(2),  151--159 (2013)

\bibitem{eigenpro}
Ma, S., Belkin, M.: Diving into the shallows: a computational perspective on
  large-scale shallow learning. Advances in Neural Information Processing
  Systems  \textbf{30} (2017)

\bibitem{ma2014}
Ma, Y., Guo, G.: Support vector machines applications, vol.~649. Springer
  (2014)

\bibitem{osuna1997}
Osuna, E., Freund, R., Girosi, F.: An improved training algorithm of support
  vector machines. In: Neural Networks for Signal Processing VII. pp. 276 --
  285 (1997)

\bibitem{qaadan2019}
Qaadan, S., Schüler, M., Glasmachers, T.: Dual {SVM} training on a budget. In:
  Proceedings of the 8th International Conference on Pattern Recognition
  Applications and Methods. SCITEPRESS - Science and Technology Publications
  (2019)

\bibitem{rahimi2008random}
Rahimi, A., Recht, B.: Random features for large-scale kernel machines. In:
  Advances in neural information processing systems. pp. 1177--1184 (2008)

\bibitem{pegasos}
Shalev-Shwartz, S., Singer, Y., Srebro, N.: Pegasos: Primal estimated
  sub-gradient solver for {SVM}. In: Proceedings of the 24th International
  Conference on Machine Learning. pp. 807--814 (2007)

\bibitem{steinwart2003sparseness}
Steinwart, I.: Sparseness of support vector machines. Journal of Machine
  Learning Research  \textbf{4},  1071--1105 (2003)

\bibitem{steinwart2011training}
Steinwart, I., Hush, D., Scovel, C.: Training {SVMs} without offset. Journal of
  Machine Learning Research  \textbf{12}(Jan),  141--202 (2011)

\bibitem{tsang2005core}
Tsang, I.W., Kwok, J.T., Cheung, P.M., Cristianini, N.: Core vector machines:
  Fast {SVM} training on very large data sets. Journal of Machine Learning
  Research  \textbf{6}(4) (2005)

\bibitem{wang2012breaking}
Wang, Z., Crammer, K., Vucetic, S.: Breaking the curse of kernelization:
  Budgeted stochastic gradient descent for large-scale {SVM} training. J. Mach.
  Learn. Res.  \textbf{13}(1),  3103--3131 (2012)

\bibitem{thundersvm}
Wen, Z., Shi, J., Li, Q., He, B., Chen, J.: {ThunderSVM}: A fast {SVM} library
  on {GPUs} and {CPUs}. The Journal of Machine Learning Research
  \textbf{19}(1),  797--801 (2018)

\bibitem{yang2012nystrom}
Yang, T., Li, Y.F., Mahdavi, M., Jin, R., Zhou, Z.H.: Nystr{\"o}m method vs.\
  random fourier features: A theoretical and empirical comparison. In: Advances
  in neural information processing systems. pp. 476--484 (2012)

\bibitem{llsvm}
Zhang, K., Lan, L., Wang, Z., Moerchen, F.: Scaling up kernel {SVM} on limited
  resources: A low-rank linearization approach. In: Artificial intelligence and
  statistics. pp. 1425--1434. PMLR (2012)

\end{thebibliography}
\end{document}